\documentclass{neus2025}
\usepackage{booktabs}

\title[Scenario Generation for ADS Testing using Failure Records]{Scenario Generation for Testing of Autonomous Driving Systems Using Real-World Failure Records}
\usepackage{times}
\usepackage{tcolorbox}
\tcbuselibrary{breakable}
\usepackage{listings}

\usepackage{appendix}
\usepackage{siunitx}
\usepackage{caption}
\usepackage{subcaption}
\usepackage{adjustbox}
\usepackage[table]{xcolor}

\coltauthor{\Name{Anjali Parashar} \Email{anjalip@mit.edu}\\
 \Name{Chuchu Fan} \Email{chuchu@mit.edu}\\
 \addr Massachusetts Institute of Technology, Cambridge, MA}

\definecolor{SigA}{HTML}{DDEEFF} 
\definecolor{SigB}{HTML}{E6F5E6} 
\definecolor{SigC}{HTML}{FFF2CC} 
\definecolor{SigD}{HTML}{FCE5CD} 
\definecolor{SigE}{HTML}{F4CCCC} 
\definecolor{SigF}{HTML}{D9EAD3} 
\definecolor{SigG}{HTML}{EAD1DC} 
\definecolor{SigH}{HTML}{D0E0E3} 
\definecolor{SigI}{HTML}{CFE2F3} 
\definecolor{SigJ}{HTML}{D9D2E9} 
\definecolor{SigK}{HTML}{E2EFDA} 

\newcommand{\sigcell}[2]{\cellcolor{#1}\textbf{#2}}

\begin{document}

\maketitle

\begin{abstract}
To ensure safe on-road behavior, pre-deployment testing and failure discovery of Autonomous Driving Systems (ADS) is crucial. Present day simulation based testing methods focus largely on mathematical models for efficient search of optimal scenarios, assuming a fixed scenario representation. On the other hand, real-world testing involves substantial manual effort to design scenario templates for testing. These templates represent distinct failure scenarios consisting of pre-deployment vehicle movements, map types, etc. Historical failure records for ADS are a reliable source of real-world failure conditions, which can be used for scenario generation. In this work, we propose a scenario generation pipeline using categorical and contextual information available from historical records in natural language format. Our approach consists of modular LLM based synthetic scenario generation, compatible with the testing constraints of a given system. We successfully apply our method to generate a diverse set of scenarios for testing autonomous navigation on Metadrive simulator using the NHTSA ADS crash records. Our approach results in accurate and diverse scenario generation with a combination of 4 road types, 3 non ego vehicle movement types, including on road anomalies in the form of working zones. Generated scenarios align with the provided testing conditions, and reveals interesting failures of the system within a limited testing budget of 20 scenarios. Code is available at \url{https://github.com/anjaliParashar/crash2scenario}. 
\end{abstract}


\section{Introduction}

Failure discovery, or falsification, of Autonomous Driving Systems (ADS) is a key step in ensuring on-road safety of vehicle occupants and surrounding entities on road \citep{najm2007pre,10.1007/978-981-96-7956-0_7,szenasi2021analysis}. Present day methods for real-world ADS falsification focus on testing system performance over a suite of manually designed scenarios \citep{berger2015large}. The goal of test suite design is to cover a wide range of system functionalities and real-world failure scenarios. Here, time and cost impose constraints on the number of scenarios that can be tested. Hence, extensive manual efforts are devoted to designing a set of \textit{scenario templates}, such as Car-to-Car Rear braking (CCRb), adult crossing the road, etc., representing a range of real world failure settings \citep{euroncap_website}. Each \textit{scenario template} is further parameterized by a small number of scenario parameters, such as vehicle speed, which are varied across a uniform range to generate a combination of scenarios for testing the vehicle performance. 

The scenario templates and parameters are chosen by experts based on historical failure data to recreate failure-relevant conditions, and test the vehicle performance across a range of operational settings. Also dubbed as combinatorial testing, this structure is followed by regulatory testing authorities such as Euro New Car Assessment Programme (NCAP), NHTSA, etc \citep{hackney1995new}. The resulting test suite is compatible with real-time testing requirements (time and cost), and provides a user-centric baseline for vehicle performance on road.

As sensor capabilities and ADS functionalities advance with engineering innovations, these hand-designed, static test suites need to be revised periodically \citep{EuroNCAP_2026_Protocols}. Additionally, the generated test suites cannot be easily adapted to stress test a specific vehicle or system. Therefore, it is possible that a vehicle crashes in real-time due to a failure that has been validated, as a potential failure may remain undiscovered in testing \citep{teambhp_ncap_overoptimisation}. 

As an alternative to static test suites, several adaptive scenario generation frameworks have emerged in simulation-based testing ~\citep{klampfl2024testing,zhu2022review,dawson2023a,koren2018adaptive,koren2021finding,ghaiGeneratingAdversarialDisturbances2021,hanselmannKINGGeneratingSafetyCritical2022a,wong2018provable,karve2026optimizing}. However, these methods assume access to a pre-defined scenario template and focus primarily on leveraging mathematical tools for efficient search of numerical scenario parameters associated to a fixed scenario template and set of parameters.

Historical ADS crash records capture both common and non-trivial real-world failures that fixed scenario templates miss \citep{NHTSA_ADAS_Crash_Report_SGO}. For example, the CCRb template only varies ego and target velocities, ignoring contextual factors like work zones that substantially increase failure likelihood. This leaves a gap in testing practice, as there is no principled way to extract scenario definitions from crash data and use them for system-specific testing. While simulation based testing approaches encourage exploration of numerical scenario parameters \citep{delecki2022we,dawson2023a}, it is also important to cover all types of failure scenarios for thorough testing.


In this work, we propose a scenario generation framework that uses data from existing crash records in a principled way, while incorporating system specific testing constraints. We construct a test suite of \textit{scenario templates} parameterized using existing categorical features available in the crash records, such as pre-crash vehicle movements, road type, etc. Crash records also provide natural language description of the incident, called \textit{narratives}, providing detailed, granular insights about failure conditions.  Recently, \cite{leung2025road} successfully demonstrated scene synthesis in autonomous driving using LLMs. We adopt this concept to transform the narratives of crashes to a system specific scenarios using LLM assisted scenario generation pipeline. Our proposed approach can be easily used for initial scenario generation, over which low-level testing methods such as Bayesian Optimization \citep{parashar2024failure}, MCMC sampling \citep{parashar2024learning,delecki2022we}, and gradient based optimization \citep{dawson2023a} can be deployed to further iterate and optimize over numerical parameters. 


We validate our end-to-end scenario generation pipeline by designing scenarios using NHTSA ADS crash records \citep{NHTSA_SGO_2021}. The scenarios are designed for testing \textsf{IDM policy} on \textsf{Metadrive} simulator \citep{li2022metadrive} for AV navigation and obstacle avoidance. As discussed in Section~\ref{sec:experiments}, our approach successfully transforms natural language descriptions failure scenarios, and generates them for simulation. Using the historical records, we test the system on common as well as uncommon failure conditions. Our main contributions are as follows:
\begin{itemize}
    \item We present an automated scenario generation procedure that utilizes the categorical feature based scenario template design, as well as incident-specific narrative available from real-world ADS crashes.   
    \vspace{-0.75em}
    \item We propose a method to adapt the available records to system specific test scenarios using LLM based scenario generation pipeline.
   \vspace{-0.5em}
    \item We validate our approach for system specific testing of crashes extracted from NHTSA crash records using the \textsf{Metadrive} simulator. 
\end{itemize}

\section{Related Works}
The key approaches in autonomous system testing can be divided in three categories, based on cost of testing, leading to different mathematical tools that we discuss below.


\subsection{Simulation based testing using sampling based approaches}

These methods assume access to a cheap simulation model, and construct failure discovery as a cost-guided search for a specific dynamic system, using sampling-based methods~\cite{delecki2022we,okellyScalableEndtoEndAutonomous2018a,sinha2020neural,dawson2023a,koren2018adaptive} and optimization techniques~\citep{wong2018provable,hanselmannKINGGeneratingSafetyCritical2022a}. These approaches do not incorporate the cost of test case generation, therefore, they focus on leveraging sample extensive mathematical frameworks such as MCMC sampling \citep{dawson2023a,delecki2022we}, MCTS \citep{koren2018adaptive} for efficient search space coverage, discovery of rare failures, and average system validation. Generally, scenario parameterization is a user-defined choice, and not directly informed by existing crash records, as in the case of regulated real-world testing. 
\subsection{Real-world testing using historical records}
Combinatorial testing practices for real-world testing leverage information from recreating historical crash records \citep{EuroNCAP_2026_Protocols}. Euro NCAP protocols assess for safety metrics across multiple criteria, which can be broadly divided into three categories-- focus on protection of vehicle inhabitants (adult and child) from the impact of collision, Vulnerable Road Users (VRUs), (pedestrians and bicyclists) and non-ego actors (cars). For each category of ego and non-ego agent, a variety of scenario templates are defined using a set of scenario parameters, such as speed of vehicle, relative location and motion of non-ego actor, and time-to-collision (TTC).  
The breadth of testing conditions covered by these scenarios is intended to provide a rich scenario coverage. While these approaches provide a baseline to compare between different vehicles, they do not stress-test a given vehicle to expose system specific failures. Additionally, they assume a rigid structure for designing scenario templates, rendering the overall pipeline non-transferable to simulation  testing. 

\subsection{Combined sim and real testing using multi-fidelity approaches}
 Recently, some works have considered cost aware adaptive testing using Bayesian Experimental Design (BED) based formulations, using a surrogate model and sequential sampling of scenarios \citep{parashar2025cost,sinha2024rate,parashar2024failure}. Some of these works also extend to joint testing using simulation and real world systems, by using multi-fidelity scenario generation and incorporating cost associated to each fidelity \citep{parashar2024failure}. These approaches use a mix of active learning/BED and sampling based approaches, and also make fundamental assumptions regarding pre-defined access to scenario parameterization. 

Our work is complementary to each of these approaches, as we focus fundamentally on principled scenario design using historical crash records as a realistic source of scenario template design and parameterization. set of scenario templates extracted using our method can be plugged in to any of the above approaches, since we treat system specific adaptation as a modular component of our pipeline. 




\section{Problem Statement}

\paragraph{Definitions.}
Consider an ADS, defined as $\dot{x} = f(x,\pi(o,z_s))$ with state $x \in \mathcal{X}$, with system observation $o \in \mathcal{O}$ and scenario parameters $z_s \in \mathcal{Z}$, which takes action $a \in \mathcal{A}$ based on a certain policy $\pi: \mathcal{O} \times \mathcal{Z} \to \mathcal{A}$. We assume access to the system with pre-defined policy $\pi$ that can take $z_s$ as input and generate trajectory rollouts, and evaluate system performance using metrics $y \in \mathcal{Y}$. These metrics are assumed to be user-defined, for example, TTC, minimum gap, etc.  
The subscript $s$ in $z_s \in \mathcal{S}$ denotes a \textit{scenario template}, which contains sufficient contextual information to uniquely define a collection of scenarios that can be distinguished from other scenarios in $\mathcal{S}$ using $M$ \textit{meta variables} $[s_i]_{i=1}^M$. For example, road type, ego and non-ego vehicle trajectory type, and weather are meta variables. Each unique combination of these meta variables  can be used to define a scenario template. For each $s \in \mathcal{S}$, we define a scenario $z_s \in \mathcal{Z}$ using numerical values of variables such as ego vehicle speed, initial gap between vehicles, etc.
 For example, a straight road, with ego vehicle turning left and non-ego vehicle moving straight is a scenario template $s$, for which, specifications such as length of road, speed of agents and trajectory parametrizes $z_s$.
 
The objective of this work is to propose a principled scenario design paradigm, that uses historical failure records as a reference and adequately stress tests a given system. We use these records in a principled manner as a way to mitigate the manual effort that goes in scenario design. The generated scenarios can be adapted for system specific testing, alleviating common concerns with static testing methods.
In this work, we use ADS crash report format available in NHTSA records for constructing scenario templates. Our scenario design pipeline consists of three-step LLM based template generation to use crash data to design scenario templates for system specific testing. This can subsequently be used to sample scenarios in a cost aware manner while ensuring diversity across generated test cases, as we show in Section \ref{sec:experiments}.




\section{Scenario design using historical failure data}

Our scenario design approach consists of two main steps, (1) design of \textit{meta variables} using crash records, and (2) LLM based scenario design using crash records as inputs.  The first  step provides an example for how to use historical failure records for efficient scenario representation. The second step is modular and can be used with other failure database as well. 
\subsection{Extracting information from historical failure data}
The information from each crash in the NHTSA ADS crash records can be divided in two categories:
  \begin{itemize}
  \vspace{-0.5em}
      \item \textit{Meta variables}: Categorical variables denoting quantities of interest, such as `Pre-crash movement Crash Partner (CP)', `Pre-crash movement Self Vehicle (SV)', `Road Type', etc. The set of discrete variables that can be supported by the simulator/system are used to define \textit{meta variables}. 
      \vspace{-1em}
      \item  \textit{Narrative}: Contextual information pertaining to the scene, recorded in the form of \textit{Narrative}, which describes qualitative aspects of the ego and non-ego trajectories, and their relative location in the global frame of reference denoted by a road type. 
  \end{itemize} 
Table~\ref{tab:discrete-var} shows meta variables extracted from NHTSA ADS records. The set of meta variables selected must be compatible with scenario generation capabilities of the system/simulator. These meta variables can be used to define a scenario template in multiple ways. The minimum requirement here is that each combination of meta variables must provide a high level interpretable categorization to distinguish between different failure scenarios. For this, we directly use the categorical features from crash records. These variables have also been used in the design of ontological scenario models for AV testing in literature \citep{schuldt2017beitrag,bagschik2018ontology}. 
For example, a vehicle \textbf{proceeding straight} ($s_1$) on a \textbf{highway} ($s_2$) colliding with another vehicle \textbf{making a left turn} ($s_3$) defines a scenario template. Here, the corresponding meta variables are Pre-crash movement SV, Road Type and Pre-Crash movement CP respectively. Fig.~\ref{fig:narrative-design} shows narratives for a scenario template from meta variables: \textit{Road Type}-Intersection, \textit{SV movement}-Proceeding Straight, \textit{CP movement}-Making Left Turn, \textit{Work zone}-False. The collision reports nominally report scenario description of short term activity of SV to a collision, and therefore involve two agents, SV and CP. This can be evolved to construct scenario template including long term reactive behavior of other traffic agents. 

\begin{table}[ht]
\centering
\setlength{\tabcolsep}{4pt} 
\renewcommand{\arraystretch}{1.2} 
\begin{tabular}{l l}
\toprule
\textbf{Meta Variable} & \textbf{Permissible Values} \\
\midrule
Road type & Intersection, Circle, Highway/Street \\
Work zone & True, False \\
Pre-crash movement CP/SV & Proceeding Straight, Making Left Turn, Making Right Turn \\
\bottomrule
\end{tabular}
\caption{Meta variables adopted from crash records with permissible values. This is chosen based on simulator and scenario generation compatibility.}
\label{tab:discrete-var}
\end{table}
Our next step uses a scenario template designed using this approach alongwith crash narrative to generate system specific scenarios for testing. 

\subsection{LLM based scenario generation}\label{sec:llm-agents}
After defining a scenario template using meta variables, we use a three step process to design system specific failure scenarios. For this, we use three LLM agents sequentially, (1) paraphrasing agent, (2) scenario generation agent, (3) fine-tuning agent. The details are discussed next.
\subsubsection{Paraphrasing a narrative.}\label{sec:paraphrase}
A \textit{narrative} consists of contextually relevant as well as incident specific details such as location, time  of crash, etc., which may not be relevant for the scenario generation process. The major topological details are already captured in meta variables, and the narrative provides further contextual details for a crash, which can assist the design of specific crash scenarios. Therefore, we first paraphrase the information from a  narrative using LLM as a paraphrasing agent to retain minimum necessary details relevant to scenario generation process. 

\begin{figure}
    \centering
    \includegraphics[width=0.9\linewidth]{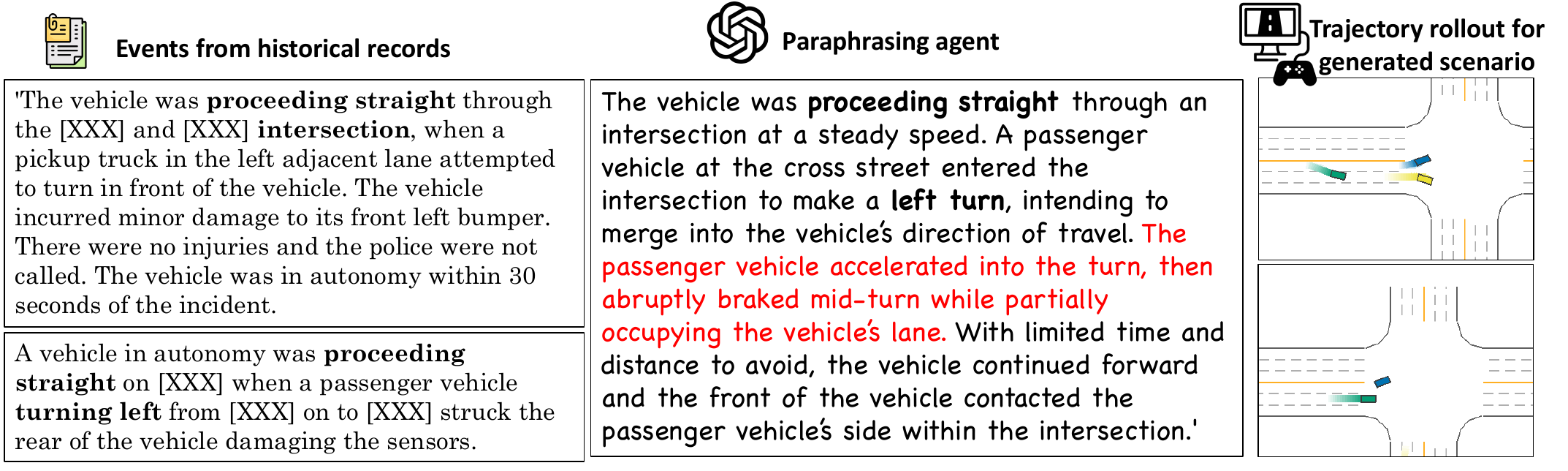}
    \caption{Extraction of Narratives for scenarios with Road Type-Intersection, CP movement- Making Left Turn, SV movement- Proceeding Straight, Work Zone-No, from historical records  (\textbf{left}). The collected data is used by the paraphrasing agent to generate a single scenario that meets design requirements. Text in red shows the system specific modification made by the agent to accommodate simulation constraints (\textbf{middle}). A scenario is generated using the proposed narrative in the simulator (\textbf{right}). }
    \label{fig:narrative-design}
\end{figure}

This step is also used to filter out scenario designs that cannot be reconstructed with the chosen system, and replace them using other alternative narrative descriptions. For example, crash records often contain collision incidents that involve rear-to-front collisions of ego vehicle (rear) with non-ego vehicle (front) due to the negligence of non-ego vehicle. However, these are not relevant for testing a specific policy deployed on our system. The paraphrasing tool assists in generating realistic, plausible alternative scenarios for such cases. Fig.~\ref{fig:narrative-design} shows an example of narrative design using the paraphrasing agent. 

\subsubsection{Scenario generation using narrative.}\label{sec:scene_gen}
The paraphrased narrative is used as an input to another LLM agent that parses the natural language description of crash, in addition to scenario template $s$, to generate a scenario $z_s$ for a possible crash. The scenario specification is encoded in the prompt, and must be directly compatible as an input to the system for testing. Appendix \ref{app:prompt-2} shows prompt used for this step. 

\begin{figure}
    \centering
    \includegraphics[width=\linewidth]{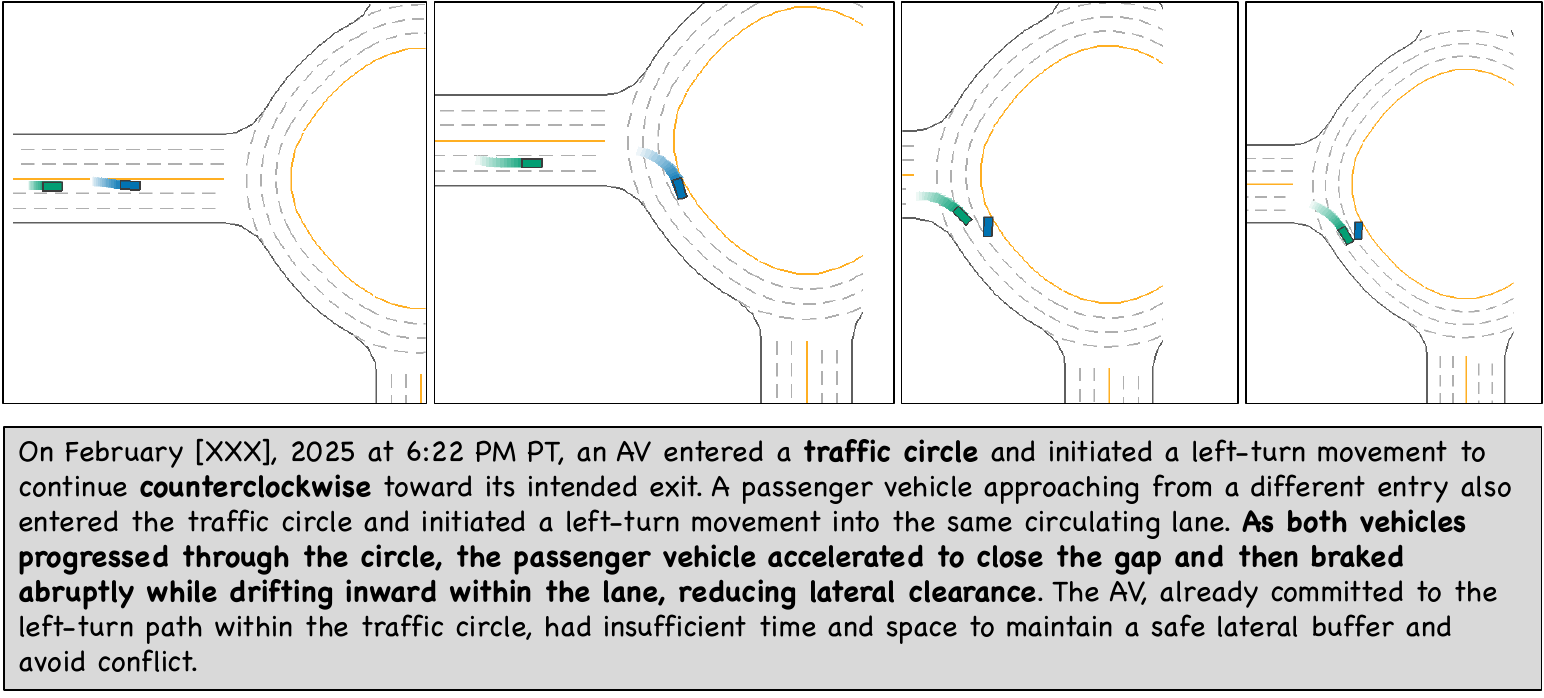}
    \caption{\textbf{Trajectory rollout for Cluster 6.} Meta variables for this example are: Road Type- Traffic Circle, Work Zone-No.  We show an exemplar LLM paraphrased narrative (Section \ref{sec:paraphrase}) at the \textbf{bottom}, and corresponding trajectory rollouts for a scenario (\textbf{top}) obtained for the  scenario generated using scenario generation (Section \ref{sec:scene_gen}).}
    \label{fig:scenario-6-narrative}
\end{figure}

The choice of specific variables and format used to define a scenario based on a given template is system and simulator specific. For example, we use \textsf{Metadrive} simulator in this work, where policies for non-ego agents can be defined to support movements such as path following using pre-available \textsf{IDM Policy}, or \textsf{Sudden Braking}, \textsf{Accelerate-then-brake} policies, which are custom policies for sudden braking, or accelerated path following followed by abrupt braking respectively. The ego agent also follows a chosen policy. In addition to movement type, attributes such as target speed of ego  $v_{e}$ and non-ego $v_n$, initial longitudinal $d_x$ and lateral gap $d_y$, map type can be used to define a scenario. Each meta variable can map to one or multiple variables. Table \ref{tab:scene-var} shows the parameters chosen to define a scenario in our experiments. Fig.~\ref{fig:scenario-6-narrative} shows a simulation for a scenario generated using the scenario generation agent using a narrative generated by the paraphrasing agent.
This step can be used with any other simulator choice such as \textsf{Scenic} \citep{fremont2023scenic}, by changing the desired schema in the prompt. 
 
\subsubsection{System specific fine-tuning.}\label{sec:fine-tune} 
\begin{figure}
    \centering
    \includegraphics[width=0.9\linewidth]{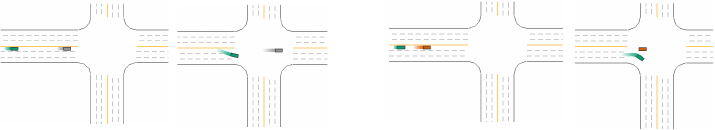}
    \caption{\textbf{Trajectory rollout for Cluster 3.} Meta variables for this scenario are: Road Type-Intersection, CP movement-Proceeding Straight, Work Zone-No. Figure shows frames corresponding to initial gap before fine tuning, \textbf{left})  and final gap for scenario fine-tuned by fine-tuning agent (\textbf{right}), showing the improvement in fatality made using LLM based fine-tuning.}
    \label{fig:fine-tuning}
\end{figure}

To refine the scenario generated by the LLM, it is helpful to steer the scenario generation process using a set of metrics $y$, that are collected at the end of each rollout. Fig.~\ref{fig:fine-tuning} shows an example of a scenario before and after fine-tuning the scenario variables using minimum gap as a metric, leading to a reduction in the minimum gap between ego and non-ego vehicles.

The scenarios generated by our paradigm can be easily integrated with existing mathematical tools for testing. As an example, we demonstrate the generation of a test suite with diverse scenarios using a clustering based approach that ensures high diversity of scenario selection within limited budget.

\section{Cost aware scenario design and testing}\label{sec:clustering}

The parameterization of crash records using meta variables and narratives can be used to cluster the records, and subsequently design synthetic scenarios that represent key characteristics of each cluster. This idea is used in limited budget data acquisition for learning, as it inherently assures coverage by only generating one scenario per cluster \citep{axiotis2024data}. 
We apply this concept to generate 20 clusters of scenarios using k-means clustering with $k=20$. 



\section{Experimental Validation}\label{sec:experiments}
We use \textsf{Metadrive} simulator with \textsf{IDM} policy for the ego agent to test our scenario generation approach using the NHTSA ADS accident records for the year 2021. First, we extract scenario templates shown in Table~\ref{tab:discrete-var}, for which permissible values are shown in Table~\ref{tab:scene-var}. Then, we use \textsf{GPT 5.2} model \citep{openai2025gpt52}  for paraphrasing narratives, converting overall information to scenario using \textsf{Metadrive} simulator, and fine-tuning the scenarios. Below we discuss each of the steps in detail. 
\subsection{Generating scenario template from crash records}
For this step, we choose meta variables from a larger set of categorical variables available in the crash records based on our setup in the simulator.
\paragraph{ Extracting meta variables from crash records.}
 From a total 2295 crashes, we eliminate data entries where 'Narrative' column is redacted, reducing the dataset to 1911 crashes. 
We restrict the movement of CP and SV to be `Proceeding Straight', `Making Left Turn' and `Making Right Turn', as these are easily reproducible in the simulator, as opposed to movements such as `Backing', or `Unknown'. 
This reduces the overall number of crashes to 235, from which we obtain the set of meta variables shown in Table~\ref{tab:discrete-var}. As discussed before, each unique combination of these meta variables defines a scenario template, leading to 25 permissible scenario templates.


As shown in Table~\ref{tab:agg_tables}, these scenario templates are not distributed uniformly by default. Infact, only 14 out of 25 scenario templates are observed in data. With 104 records out of 235, most commonly occurring scenario template corresponds to the following meta variable values: Intersection, Proceeding Straight, Proceeding Straight,False, for road-types, CP/SV pre-crash movements, and work-zone respectively. Only 9 out of 235 crashes correspond to Work Zone variable True, reflecting that crashes due to presence of Work Zone are rare. 

\begin{table*}[t]
\centering

\setlength{\tabcolsep}{5pt}
\small

\begin{minipage}{0.45\textwidth}
\centering
\textbf{(a) Road Type Distribution}

\vspace{4pt}

\begin{tabular}{l S[table-format=3.0] S[table-format=2.1]}
\toprule
Road Type & {Number of events} & {\%} \\
\midrule
Intersection      & \textbf{214} & 91.1 \\
Highway / Freeway & 20  & 8.5  \\
Traffic Circle    & \textbf{1}   & 0.4  \\
\bottomrule

\end{tabular}
\end{minipage}
\hfill
\begin{minipage}{0.5\textwidth}
\centering
\textbf{(b) CP--SV Interaction Matrix}

\vspace{4pt}

\begin{tabular}{lccc}
\toprule
CP $\backslash$ SV & Straight & Left Turn & Right Turn \\
\midrule
Straight & \textbf{120} & 26 & 16 \\
Left Turn & 36  & 9  & \textbf{1}  \\
Right Turn & 9   & 4  & 14 \\
\bottomrule
\end{tabular}

\vspace{4pt}
\end{minipage}
\caption{Crash distribution by Road Type and CP-SV pre-crash movements. We highlight the most and least frequently occurring meta variables in bold.}
\label{tab:agg_tables}
\end{table*}

\subsection{Design of scenario.}
The simulator supports five map types, \textsf{I,T,O,C,S}, representing four-way intersections, three-way intersections, roundabouts, curved roads, and straight roads. We directly use map type and ego target speed as scenario parameters. To model Crash Partners (CPs), we add two non-ego policies with irregular motion, \textsf{Sudden} and \textsf{Accelerate-then-brake}, alongside nominal \textsf{IDM}. 
Scenarios are generated using the three-stage LLM-assisted pipeline illustrated in Section~\ref{sec:llm-agents}. The crash records for scenario generation are selected using a clustering method, to ensure generation of diverse scenarios, that we briefly discuss below.



\subsection{Clustering to generate typical scenarios}\label{sec:clustering}
We cluster the collection of crashes into 20 clusters using k-means clustering using the 14 scenario templates, represented in the form of one-hot encoding, alongwith a 6-dimensional vector embedding of the narratives as features for clustering. The embedding is generated using \textsf{all-MiniLM-L6-v2} \citep{reimers2019sentence,wang2020minilm} and compressed with PCA. This representation encodes diversity of scenarios by frequency of failure, contextual details, as well as meta variables. The resulting clusters represent a collection of scenarios that are rare as well as frequent, correspond to different spatio-temporal types, and different contextual specifics encoded in the narrative.   
Table~\ref{tab:cluster_centroids_grouped} shows the meta variables corresponding to each cluster. We observet that including the narratives in clustering further adds benefit. Firstly it helps to segregate between contextually different scenarios with same scenario template. Second, it helps in segregating scenarios that do not have incident relevant information (for example, Cluster 3). Appendix~\ref{app:narrative} shows narratives generated for each cluster using the paraphrasing agent.
\begin{table}[t]
\centering
\begin{adjustbox}{max width=\linewidth}
\begin{tabular}{l*{20}{c}}
\toprule
 & \multicolumn{20}{c}{\textbf{Cluster ID (grouped by scenario template: Road, SV, CP, WZ)}} \\
\cmidrule(lr){2-21}
 & 
\sigcell{SigA}{2} &
\sigcell{SigB}{10} & \sigcell{SigB}{18} &
\sigcell{SigC}{4} &
\sigcell{SigD}{1} & \sigcell{SigD}{14} &
\sigcell{SigE}{0} & \sigcell{SigE}{8} & \sigcell{SigE}{19} &
\sigcell{SigF}{3} & \sigcell{SigF}{9} & \sigcell{SigF}{16} & \sigcell{SigF}{17} &
\sigcell{SigG}{15} &
\sigcell{SigH}{6} &
\sigcell{SigI}{7} &
\sigcell{SigJ}{5} & \sigcell{SigJ}{11} &
\sigcell{SigK}{12} & \sigcell{SigK}{13}
\\
\midrule

CP &
LT &
PS & PS &
PS &
LT & LT &
LT & LT & LT &
PS & PS & PS & PS &
RT &
LT &
LT &
PS & PS &
PS & PS
\\

SV &
RT &
LT & LT &
LT &
PS & PS &
LT & LT & LT &
PS & PS & PS & PS &
RT &
LT &
PS &
PS & PS &
PS & PS
\\

WZ &
N &
N & N &
Y &
N & N &
N & N & N &
N & N & N & N &
N &
N &
N &
N & N &
Y & Y
\\

Road &
Int &
Int & Int &
Int &
Int & Int &
Int & Int & Int &
Int & Int & Int & Int &
Int &
Circle &
Hwy &
Hwy & Hwy &
Hwy & Hwy
\\

\bottomrule
\end{tabular}
\end{adjustbox}
\caption{Cluster centroids grouped by scenario templates (same color for identical Road/SV/CP/WZ, PS: Proceeding Straight, LT: Making Left Turn, RT: Making Right Turn; Int: Intersection, Hwy: Highway/Freeway, Circle: Traffic Circle; WZ: N/Y.}
\label{tab:cluster_centroids_grouped}
\end{table}

\begin{table}[t]
\centering

\begin{adjustbox}{max width=\linewidth}
\begin{tabular}{l*{20}{c}}
\toprule
& \multicolumn{20}{c}{\textbf{Cluster ID}} \\
\cmidrule(lr){2-21}
& 0 & 1 & 2 & 3 & 4 & 5 & 6 & 7 & 8 & 9 & 10 & 11 & 12 & 13 & 14 & 15 & 16 & 17 & 18 & 19 \\
\midrule
Initial
& 20.00 & 9.04 & 30.00 & 22.79 & 1.30 & 2.08 & 3.61 & 10.02 & 30.74 & 40.14
& 28.00 & 0.10 & 1.37 & 0.13 & 20.00 & 4.31 & 34.32 & 34.00 & 11.10 & 36.26 \\
Final
& 6.53 & 6.03 & 9.02 & 5.34 & - & 0.12 & 8.01 & 10.02 & 14.01 & 12.01
& 10.02 & - & - & - & 8.02 & 3.64 & 12.01 & 12.01 & 6.09 & 18.01 \\
Diff.
& -67.33 & -33.30 & -69.93 & -76.56 & - & -94.31 & 121.73 & 0.00 & -54.42 & -70.08
& -64.22 & - & - & - & -59.90 & -15.37 & -65.01 & -64.68 & -45.15 & -50.34 \\
\bottomrule
\end{tabular}
\end{adjustbox}
\caption{Initial/Final Minimum Gap and Percentage Difference by Cluster. Clusters for which collision happens for intial scenario are not sent for fine-tuning.}
\label{tab:init-final}
\end{table}

\section{Discussion}

Our main contribution with this work is to present an automated scenario generation paradigm that uses real world records to construct realistic scenarios for testing. 
We evaluate the overall merit of our approach in testing the given policy using three key questions. Appendix~\ref{app:discuss} provides further discussion, assessing each component of our pipeline.

\textbf{(Q1) How accurate is scenario generation by our pipeline?}
For the 20 clusters considered, 17 out 20 scenarios match the scenario template meta variables completely. For 3 scenarios, one of the meta variables is not accurately represented (Cluster 6 and 18- CP movement direction is incorrect, Cluster 11-Work Zone present in the scenario but should be False). We also find the generated CP movements match the details of the narrative accurately (Fig~\ref{fig:scenario-6-narrative}). The accuracy of match and difficulty of scenario can be further optimized using multiple steps of the fine-tuning agent (Section~\ref{sec:fine-tune}). Appendix~\ref{app:prompt-3} shows the fine-tuning prompt. Table~\ref{tab:init-final} shows that the minimum gap reduces substantially using a single iteration of fine-tuning for 14 out of 16 scenarios that need fine-tuning.


\textbf{(Q2) Can this approach be effectively used for testing of autonomous vehicles?} We successfully generate interesting scenarios of collisions as well as near collisions, as evident from the minimum gap shown for all 20 scenarios in Table~\ref{tab:init-final}. We use clustering as a way to segregate the available data into diverse groups, as shown in Table~\ref{tab:cluster_centroids_grouped}. The clustering captures 11 out of the 14 scenario templates present naturally in historical records.

\textbf{(Q3) What do we learn about the underlying system using the generated tests?}
\begin{figure}[h]
    \centering
    \includegraphics[width=0.8\linewidth]{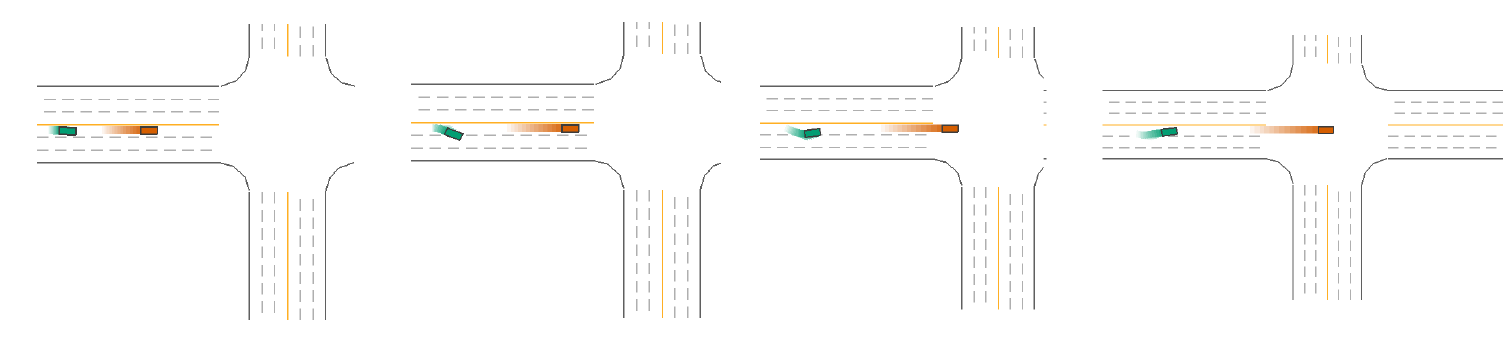}
    \caption{\textbf{Trajectory rollout for Cluster 2.} Meta variables for this scenario: CP movement-Proceeding Straight, Road Type- Intersection, Work Zone-No. While SV does not crash and maintains a safe distance from CP, SV shows oscillatory movement to avoid crash at the beginning, despite being at a sufficient distance from CP.}
    \label{fig:scenario-2-wiggle}
\end{figure}

The paraphrasing agent and scenario design agent adapt to system specific requirements. The generated scenarios reveal interesting details. We observe that for CP movement-Proceeding Straight, the \textsf{IDM policy}  ensures that the vehicle maintains a safe distance, and avoids crashes. However, this results in early braking and oscillatory movement (Fig.~\ref{fig:scenario-2-wiggle}), which is undesirable on-road. We observe that the policy is not as reactive to smaller obstacles such as cones, (Fig.~\ref{fig:scenario-work-zone}), as it is to larger vehicles, and collides often with the cones when Work Zone is set to True. While head-on collision in straight motion is avoided successfully, the policy crashes in lateral motion (Fig.~\ref{fig:scenario-6-narrative}), leading to side collisions.  

\section{Limitations \& Future Work}
In this work, we propose an automated scenario generation tool for extracting information from historical records to generate diverse testing scenarios, that can be applied to a given system. This alleviates the need to handcraft or manually design the scenario parameters. This can be easily combined with existing sampling or optimization tools for further discovery of worst case scenarios. Additionally, LLMs can be used to create synthetic failure narratives in the absence of real world information, which can be used to adaptively explore the space of scenario templates, which has not been explored in literature so far, and is a direction for future work. 

The current limitation of the work stems from a lack of unified definition for concepts such as coverage in the domain of testing \& validation, leading to non-unique ways in which diverse generation of scenarios can be pursued. Additionally, we observe that fixing a given policy gives very limited control to replicate the ego vehicle movement, thereby restricting complete replication of desired scenarios. We also aim to extend this work to the generation of visual conditions for testing of vision based policies in ADS.

\bibliography{main}

@article{szenasi2021analysis,
  title={Analysis of historical road accident data supporting autonomous vehicle control strategies},
  author={Sz{\'e}n{\'a}si, S{\'a}ndor},
  journal={PeerJ Computer Science},
  volume={7},
  pages={e399},
  year={2021},
  publisher={PeerJ Inc.}
}

@article{fremont2023scenic,
  title={Scenic: a language for scenario specification and data generation},
  author={Fremont, Daniel J and Kim, Edward and Dreossi, Tommaso and Ghosh, Shromona and Yue, Xiangyu and Sangiovanni-Vincentelli, Alberto L and Seshia, Sanjit A},
  journal={Machine Learning},
  volume={112},
  number={10},
  pages={3805--3849},
  year={2023},
  publisher={Springer}
}

@article{li2022metadrive,
  title={Metadrive: Composing diverse driving scenarios for generalizable reinforcement learning},
  author={Li, Quanyi and Peng, Zhenghao and Feng, Lan and Zhang, Qihang and Xue, Zhenghai and Zhou, Bolei},
  journal={IEEE transactions on pattern analysis and machine intelligence},
  volume={45},
  number={3},
  pages={3461--3475},
  year={2022},
  publisher={IEEE}
}

@misc{openai2025gpt52,
  author = {{OpenAI}},
  title = {GPT-5.2 Model Documentation},
  year = {2025},
  howpublished = {\url{https://platform.openai.com/docs/models}},
}

@inproceedings{leung2025road,
  title={From Road to Code: Neuro-Symbolic Program Synthesis for Autonomous Driving Scene Translation and Analysis},
  author={Leung, Johnathan and Tong, Guansen and Duggirala, Parasara Sridhar and Chakravarthula, Praneeth},
  booktitle={International Conference on Neuro-symbolic Systems},
  pages={331--351},
  year={2025},
  organization={PMLR}
}

@misc{NHTSA_ADAS_Crash_Report_SGO,
  author       = {{National Highway Traffic Safety Administration}},
  title        = {Standing General Order on Crash Reporting for vehicles equipped with Automated Driving Systems and Level 2 Advanced Driver Assistance Systems},
  howpublished = {Regulatory guidance, U.S. Department of Transportation},
  note         = {\url{https://www.nhtsa.gov/laws-regulations/standing-general-order-crash-reporting}},
  year         = {2021},
  organization = {National Highway Traffic Safety Administration},
}

@techreport{hackney1995new,
  title={The new car assessment program: five star rating system and vehicle safety performance characteristics},
  author={Hackney, James R and Kahane, Charles J},
  year={1995},
  institution={SAE Technical Paper}
}

@misc{EuroNCAP_2026_Protocols,
  author       = {{Euro NCAP}},
  title        = {Euro NCAP 2026 Protocols},
  howpublished = {Online},
  year         = {2025},
  url          = {https://www.euroncap.com/en/for-engineers/protocols/2026-protocols/},
}

@article{karve2026optimizing,
  title={Optimizing Virtual Scenario Testing for Autonomous Vehicles with Large-Scale Scenario Generation},
  author={Karve, Omkar and Saurav, Saket and Purwar, Prabhanshu},
  journal={SAE Technical Paper},
  pages={01--0050},
  year={2026},
  publisher={SAE International}
}

@article{wang2020minilm,
  title={Minilm: Deep self-attention distillation for task-agnostic compression of pre-trained transformers},
  author={Wang, Wenhui and Wei, Furu and Dong, Li and Bao, Hangbo and Yang, Nan and Zhou, Ming},
  journal={Advances in neural information processing systems},
  volume={33},
  pages={5776--5788},
  year={2020}
}

@inproceedings{reimers2019sentence,
  title={Sentence-bert: Sentence embeddings using siamese bert-networks},
  author={Reimers, Nils and Gurevych, Iryna},
  booktitle={Proceedings of the 2019 conference on empirical methods in natural language processing and the 9th international joint conference on natural language processing (EMNLP-IJCNLP)},
  pages={3982--3992},
  year={2019}
}

@article{axiotis2024data,
  title={Data-efficient learning via clustering-based sensitivity sampling: Foundation models and beyond},
  author={Axiotis, Kyriakos and Cohen-Addad, Vincent and Henzinger, Monika and Jerome, Sammy and Mirrokni, Vahab and Saulpic, David and Woodruff, David and Wunder, Michael},
  journal={arXiv preprint arXiv:2402.17327},
  year={2024}
}

@article{sinha2024rate,
  title={Rate-Informed Discovery via Bayesian Adaptive Multifidelity Sampling},
  author={Sinha, Aman and Nikdel, Payam and Paul, Supratik and Whiteson, Shimon},
  journal={arXiv preprint arXiv:2411.17826},
  year={2024}
}

@inproceedings{parashar2025cost,
  title={Cost-aware Discovery of Contextual Failures using Bayesian Active Learning},
  author={Parashar, Anjali and Zhang, Joseph and Li, Yingke and Fan, Chuchu},
  booktitle={9th Annual Conference on Robot Learning},
  year={2025}
}

@misc{NHTSA_SGO_2021,
  author       = {{National Highway Traffic Safety Administration (NHTSA)}},
  title        = {Standing General Order on Crash Reporting},
  year         = {2021},
  howpublished = {\url{https://www.nhtsa.gov/laws-regulations/standing-general-order-crash-reporting}},
}

@misc{teambhp_ncap_overoptimisation,
  author = {{Team-BHP Forum}},
  title  = {German crash test reveals NCAP over-optimisation can cause cars to protect worse at lower speeds},
  year   = {2026},
  url    = {https://www.team-bhp.com/forum/road-safety/305252-german-crash-test-reveals-ncap-over-optimisation-can-cause-cars-protect-worse-lower-speeds.html},
}

@inproceedings{berger2015large,
  title={Large-Scale Evaluation of an Active Safety Algorithm with EuroNCAP and US NCAP Scenarios in a Virtual Test Environment--An Industrial Case Study},
  author={Berger, Christian and Block, Delf and Hons, Christian and K{\"u}hnel, Stefan and Leschke, Andr{\'e} and Plotnikov, Dimitri and Rumpe, Bernhard},
  booktitle={2015 IEEE 18th International Conference on Intelligent Transportation Systems},
  pages={2280--2286},
  year={2015},
  organization={IEEE}
}

@misc{euroncap_website,
  author       = {{Euro NCAP}},
  title        = {The European New Car Assessment Programme},
  howpublished = {\url{https://www.euroncap.com/en}},

}

@inproceedings{10.1007/978-981-96-7956-0_7,
author = {Kulicki, Piotr and Trypuz, Robert},
title = {Ontology of Autonomous Driving as a Tool for Argumentation on Responsibility},
year = {2025},
isbn = {978-981-96-7955-3},
publisher = {Springer-Verlag},
address = {Berlin, Heidelberg},
doi = {10.1007/978-981-96-7956-0_7},
booktitle = {Logic and Argumentation: 6th International Conference, CLAR 2025, Taiyuan, China, June 14–16, 2025, Proceedings},
pages = {104–120},
numpages = {17},
location = {Taiyuan, China}
}

@techreport{najm2007pre,
  title={Pre-crash scenario typology for crash avoidance research},
  author={Najm, Wassim G and Smith, John D and Yanagisawa, Mikio and others},
  year={2007},
  institution={United States. Department of Transportation. National Highway Traffic Safety~…}
}

@article{klampfl2024testing,
  title={Testing ADAS/ADS--from critical scenarios to automated testing oracles},
  author={Klampfl, Lorenz and Kl{\"u}ck, Florian and Nica, Mihai and Tao, Jianbo and Wotawa, Franz},
  journal={e+ i Elektrotechnik und Informationstechnik},
  volume={141},
  number={6},
  pages={392--399},
  year={2024},
  publisher={Springer}
}

@article{zhu2022review,
  title={Review on functional testing scenario library generation for connected and automated vehicles},
  author={Zhu, Yu and Wang, Jian and Meng, Fanqiang and Liu, Tongtao},
  journal={Sensors},
  volume={22},
  number={20},
  pages={7735},
  year={2022},
  publisher={MDPI}
}

@inproceedings{koren2018adaptive,
  title={Adaptive stress testing for autonomous vehicles},
  author={Koren, Mark and Alsaif, Saud and Lee, Ritchie and Kochenderfer, Mykel J},
  booktitle={2018 IEEE Intelligent Vehicles Symposium (IV)},
  pages={1--7},
  year={2018},
  organization={IEEE}
}

@inproceedings{koren2021finding,
  title={Finding failures in high-fidelity simulation using adaptive stress testing and the backward algorithm},
  author={Koren, Mark and Nassar, Ahmed and Kochenderfer, Mykel J},
  booktitle={2021 IEEE/RSJ International Conference on Intelligent Robots and Systems (IROS)},
  pages={5944--5949},
  year={2021},
  organization={IEEE}
}

@inproceedings{dawson2023a,
title={A {B}ayesian approach to breaking things: efficiently predicting and repairing failure modes via sampling},
author={Charles Dawson and Chuchu Fan},
booktitle={7th Annual Conference on Robot Learning},
year={2023},
url={https://openreview.net/forum?id=fNLBmtyBiC}
}

@inproceedings{wong2018provable,
  title={Provable defenses against adversarial examples via the convex outer adversarial polytope},
  author={Wong, Eric and Kolter, Zico},
  booktitle={International conference on machine learning},
  pages={5286--5295},
  year={2018},
  organization={PMLR}
}

@article{sinha2020neural,
  title={Neural bridge sampling for evaluating safety-critical autonomous systems},
  author={Sinha, Aman and O'Kelly, Matthew and Tedrake, Russ and Duchi, John C},
  journal={Advances in Neural Information Processing Systems},
  volume={33},
  pages={6402--6416},
  year={2020}
}

@inproceedings{ghaiGeneratingAdversarialDisturbances2021,
  title = {Generating {{Adversarial Disturbances}} for {{Controller Verification}}},
  booktitle = {Proceedings of the 3rd {{Conference}} on {{Learning}} for {{Dynamics}} and {{Control}}},
  author = {Ghai, Udaya and Snyder, David and Majumdar, Anirudha and Hazan, Elad},
  year = {2021},
  month = may,
  pages = {1192--1204},
  publisher = {PMLR},
  issn = {2640-3498},
  urldate = {2023-03-19},
  langid = {english}
}

@inproceedings{hanselmannKINGGeneratingSafetyCritical2022a,
  title = {{{KING}}: {{Generating Safety-Critical Driving Scenarios}} for~{{Robust Imitation}} via~{{Kinematics Gradients}}},
  shorttitle = {{{KING}}},
  booktitle = {Computer {{Vision}} -- {{ECCV}} 2022: 17th {{European Conference}}, {{Tel Aviv}}, {{Israel}}, {{October}} 23--27, 2022, {{Proceedings}}, {{Part XXXVIII}}},
  author = {Hanselmann, Niklas and Renz, Katrin and Chitta, Kashyap and Bhattacharyya, Apratim and Geiger, Andreas},
  year = {2022},
  month = oct,
  pages = {335--352},
  publisher = {Springer-Verlag},
  address = {Berlin, Heidelberg},
  doi = {10.1007/978-3-031-19839-7_20},
  urldate = {2023-09-15},
  isbn = {978-3-031-19838-0}
}

@inproceedings{okellyScalableEndtoEndAutonomous2018a,
  title = {Scalable {{End-to-End Autonomous Vehicle Testing}} via {{Rare-event Simulation}}},
  booktitle = {Advances in {{Neural Information Processing Systems}}},
  author = {O' Kelly, Matthew and Sinha, Aman and Namkoong, Hongseok and Tedrake, Russ and Duchi, John C},
  year = {2018},
  volume = {31},
  publisher = {Curran Associates, Inc.},
  urldate = {2023-03-14}
}

@inproceedings{delecki2022we,
  title={How do we fail? stress testing perception in autonomous vehicles},
  author={Delecki, Harrison and Itkina, Masha and Lange, Bernard and Senanayake, Ransalu and Kochenderfer, Mykel J},
  booktitle={2022 IEEE/RSJ International Conference on Intelligent Robots and Systems (IROS)},
  pages={5139--5146},
  year={2022},
  organization={IEEE}
}

@article{parashar2024learning,
  title={Learning-based Bayesian Inference for Testing of Autonomous Systems},
  author={Parashar, Anjali and Yin, Ji and Dawson, Charles and Tsiotras, Panagiotis and Fan, Chuchu},
  journal={IEEE Robotics and Automation Letters},
  year={2024},
  publisher={IEEE}
}

@inproceedings{parashar2024failure,
  title={Failure Prediction from Few Expert Demonstrations},
  author={Parashar, Anjali and Garg, Kunal and Zhang, Joseph and Fan, Chuchu},
  year={2024},
  booktitle={NeurIPS 2024 Workshop on Bayesian Decision-making and Uncertainty}
}

@phdthesis{schuldt2017beitrag,
  title={Ein Beitrag f{\"u}r den methodischen Test von automatisierten Fahrfunktionen mit Hilfe von virtuellen Umgebungen},
  author={Schuldt, Fabian},
  year={2017}
}

@inproceedings{bagschik2018ontology,
  title={Ontology based scene creation for the development of automated vehicles},
  author={Bagschik, Gerrit and Menzel, Till and Maurer, Markus},
  booktitle={2018 IEEE intelligent vehicles symposium (IV)},
  pages={1813--1820},
  year={2018},
  organization={IEEE}
}
\newpage
\appendix



\section{Discussion of results}\label{app:discuss}
\begin{figure}
    \centering
    \includegraphics[width=0.8\linewidth]{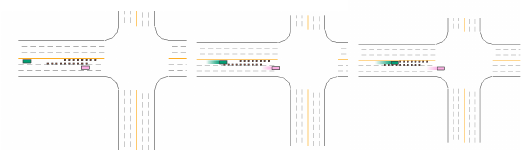}
    \includegraphics[width=0.8\linewidth]{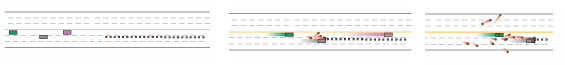}
    \caption{\textbf{Trajectory rollout for Cluster 4 \& 13.} Meta variables for this scenario: CP movement-Proceeding Straight, Road Type- Intersection (Cluster 4), Highway/Freeway (Cluster 13), Work Zone-Yes.}
    \label{fig:scenario-work-zone}
\end{figure}
\textbf{(Q4) Why do we need a separate paraphrasing agent to generate synthetic narratives?}
Firstly, narratives may not be available for all kinds of scenario templates. In such cases,  the LLM-based paraphrasing agent (Section~\ref{sec:paraphrase}) is useful in generating narratives based on scenario templates due to its generalizable performance. For example, for Cluster-3 (LLM generated narrative shown in Appendix~\ref{app:narrative}), the narrative records provide no incident relevant information, and the paraphrasing agent creates a realistic description of crash. 
Second, this helps in incorporating system specific constraints in downstream scenario design by revising the narrative such that it is feasible to recreate for a given simulator/system. For example, Fig.~\ref{fig:narrative-design} shows text in red, introduced by the paraphrasing agent using the information on types of policies that can be used to replicate non-ego agent (CP) behavior. This information can be provided in the prompt (Appendix~\ref{app:prompt-1}).

\textbf{(Q5) How useful are historical crash records in generating realistic testing conditions? }

Unlike large scale on-road self-driving datasets, historical records encode concise records focused only on on-road accidents. This mitigates the need for data-processing to extract failures. These records are used by regulatory authorities for scenario design for testing, however this tends to be a largely manual effort. One of the key contributions of our work is to automate this process, where LLM based agents are used to take system constraints into account and design scenarios for testing based on provided records. The historical records provide a natural way to categorize available data, which we leverage in our scenario template construction (Table~\ref{tab:discrete-var}). 

\section{Non-ego policy hyperparameter}\label{app:policies}
In addition to target speeds of each CP, we also provide initial position of each CP relative to SV, using lateral and longitudinal gap, ($\Delta_t,\Delta_s$), lane offset, $\{+1,0,-1\}$, which corresponds to right lane, same lane, and left lane. We also provide an option to initialize the vehicle in the same or opposite direction, leading to rear or head-on collisions. We further enable background traffic and control traffic density as a parameter between 0 and 1. Higher background traffic increases chances of collisions, but also increases randomization in scenario generation. These values are summarized in Table~\ref{tab:scene-var} and can be directly provided as scenario input to the \textsf{Metadrive} simulator.


\section{Clustering scenario details}\label{app:narrative}

\begin{tcolorbox}[colback=gray!10, colframe=gray!50, boxrule=0.5pt]

\textbf{Cluster-0.}

Road: Intersection, SV: Making Left Turn, CP:Making Left Turn, Work Zone:No 

\textbf{Narrative generated.}
The vehicle was proceeding straight through an intersection at a steady speed. A passenger vehicle at the cross street entered the intersection to make a left turn, intending to merge into the vehicle’s direction of travel. The passenger vehicle accelerated into the turn, then abruptly braked mid-turn while partially occupying the vehicle’s lane. With limited time and distance to avoid, the vehicle continued forward and the front of the vehicle contacted the passenger vehicle’s side within the intersection.
\end{tcolorbox}

\begin{tcolorbox}[colback=gray!10, colframe=gray!50, boxrule=0.5pt]
\label{box:cluster1}
\textbf{Cluster-1.}

Road: Intersection, SV: Proceeding Straight, CP:Making Left Turn, Work Zone:No 

\textbf{Narrative generated.}

The vehicle was proceeding straight through an intersection on the main road. A passenger vehicle on the cross street entered the intersection and began a left turn to merge onto the vehicle’s roadway in the same direction of travel. The turning vehicle accelerated into the near lane and encroached into the vehicle’s lane without yielding, reducing the available gap. The vehicle braked but was unable to avoid contact, striking the turning vehicle near its left-rear quarter as it completed the left turn.
\end{tcolorbox}

\begin{tcolorbox}[colback=gray!10, colframe=gray!50, boxrule=0.5pt]

\textbf{Cluster-2.}

Road: Intersection, SV: Making Right Turn, CP:Making Left Turn, Work Zone:No 

\textbf{Narrative generated.}

On April [XXX], 2025 at 3:12 PM local time, the self-driving vehicle (SDV) approached a standard four-way signalized intersection in the right-turn lane intending to make a right turn onto the cross street. A crash-partner passenger vehicle was traveling proceeding straight through the intersection on the cross street (from the SDV’s left to right). As the SDV began its right turn at low speed, the crash-partner vehicle continued straight at a moderate speed and closed the gap faster than expected. The SDV entered the near lane of the cross street while the crash-partner vehicle was still within the conflict zone, resulting in an angle impact in the intersection/near-corner area (front-right area of the SDV contacting the side/front-quarter of the crash-partner vehicle). No work zone activity was present.
\end{tcolorbox}

\begin{tcolorbox}[colback=gray!10, colframe=gray!50, boxrule=0.5pt]
\label{box:cluster3}
\textbf{Cluster-3.}

Road: Intersection, SV: Proceeding Straight CP: Proceeding Straight, Work Zone:No 

\textbf{Narrative generated.}

On an evening in 2023, the self-driving vehicle (“Ego”) was traveling straight through a signalized four-way intersection on a steady approach speed with the traffic signal indicating a permissive phase for Ego to proceed. A passenger vehicle (“Crash Partner”) approached the same intersection from the cross street and continued straight into the intersection against its signal. As the Crash Partner entered the intersection, it briefly accelerated and then abruptly braked while still within the conflict zone, reducing the available gap. Ego initiated braking but, due to the late incursion and the Crash Partner’s sudden deceleration in the intersection, Ego was unable to fully avoid the crossing vehicle and the vehicles collided in the intersection in a right-angle (side-impact) configuration.
\end{tcolorbox}

\begin{tcolorbox}[colback=gray!10, colframe=gray!50, boxrule=0.5pt]

\textbf{Cluster-4.}

Road: Intersection, SV: Making Left Turn CP: Proceeding Straight, Work Zone: Yes 

\textbf{Narrative generated.}

A vehicle (Vehicle 1) in autonomous mode approached a coned work-zone intersection and prepared to make a left turn. The temporary cone pattern narrowed and shifted the travel lanes through the intersection, placing Vehicle 1’s turn path closer to the cross-traffic lane than usual. After slowing, Vehicle 1 began the left turn at low speed. A passenger vehicle (Vehicle 2) traveling straight through the intersection in the adjacent through lane continued forward at a steady speed. As Vehicle 1 entered the intersection and crossed the straight-through path, Vehicle 2 reached the conflict point and struck Vehicle 1 in the intersection, resulting in an angle/side-impact collision.
\end{tcolorbox}

\begin{tcolorbox}[colback=gray!10, colframe=gray!50, boxrule=0.5pt]

\textbf{Cluster-5.}

Road: Highway / Freeway, SV: Proceeding Straight CP: Proceeding Straight, Work Zone: No 

\textbf{Narrative generated.}

On June [XXX], 2024 at 7:42 AM PT, the ego vehicle was traveling eastbound on a multi-lane freeway, proceeding straight in the center lane at highway speed. A crash partner vehicle was proceeding straight in the adjacent right lane at a similar speed. As both vehicles continued forward, the crash partner gradually drifted left within its lane and then encroached over the lane line into the ego vehicle’s lane without a turn or lane-change maneuver. The ego vehicle applied braking to maintain a safe separation, but the crash partner’s continued encroachment resulted in contact between the ego vehicle’s left-front corner and the crash partner’s right-rear quarter panel while both vehicles were still moving straight.
\end{tcolorbox}

\begin{tcolorbox}[colback=gray!10, colframe=gray!50, boxrule=0.5pt]

\textbf{Cluster-6.}

Road: Traffic Circle, SV: Making Left Turn, CP: Making Left Turn, Work Zone: No 

\textbf{Narrative generated.}

On February [XXX], 2025 at 6:22 PM PT, an AV entered a traffic circle and initiated a left-turn movement to continue counterclockwise toward its intended exit. A passenger vehicle approaching from a different entry also entered the traffic circle and initiated a left-turn movement into the same circulating lane. As both vehicles progressed through the circle, the passenger vehicle accelerated to close the gap and then braked abruptly while drifting inward within the lane, reducing lateral clearance. The AV, already committed to the left-turn path within the traffic circle, had insufficient time and space to maintain a safe lateral buffer and avoid conflict. The right-front of the passenger vehicle contacted the left-front quarter of the AV within the circulating portion of the traffic circle.
\end{tcolorbox}

\begin{tcolorbox}[colback=gray!10, colframe=gray!50, boxrule=0.5pt]

\textbf{Cluster-7.}
Road: Highway / Freeway, SV: Proceeding Straight, CP: Making Left Turn, Work Zone: No 

\textbf{Narrative generated.}
Pre-crash movement through telematics data showed the self vehicle proceeding straight in the rightmost travel lane on a highway at a steady speed. A crash partner vehicle was initially positioned on the right shoulder ahead of the self vehicle, then accelerated into the roadway and initiated a left turn across the travel lanes toward a median crossover/opening. The self vehicle applied braking while continuing straight, but the crash partner continued the left-turn maneuver into the self vehicle’s lane, resulting in a front-corner-to-side impact as the self vehicle struck the crash partner during the crossing movement.
\end{tcolorbox}

\begin{tcolorbox}[colback=gray!10, colframe=gray!50, boxrule=0.5pt]

\textbf{Cluster-8.}

Road: Intersection, SV:  Making Left Turn, CP: Making Left Turn, Work Zone: No 

\textbf{Narrative generated.}

At approximately 3:18 p.m. on a clear weekday afternoon, the self-driving vehicle approached a signal-controlled intersection in the southbound lane and initiated a permitted left turn to travel eastbound. At the same time, a crash partner vehicle approached the same intersection from the westbound approach and began a right turn to travel northbound. As the self-driving vehicle entered the intersection and continued its left-turn path, the crash partner vehicle rolled forward and then accelerated into its right turn, entering the intersection later than expected. The vehicles converged in the intersection turn area, and the front of the crash partner vehicle struck the right side of the self-driving vehicle during the left turn.
\end{tcolorbox}

\begin{tcolorbox}[colback=gray!10, colframe=gray!50, boxrule=0.5pt]

\textbf{Cluster-9.}

Road: Intersection, SV: Proceeding Straight, CP: Proceeding Straight, Work Zone: No 

\textbf{Narrative generated.}

On March [XXX], 2024 at 4:22 PM PT, the ego vehicle was proceeding straight northbound through a signalized intersection at a steady speed. A crash partner vehicle was proceeding straight eastbound on the cross street toward the same intersection. As the ego vehicle entered the intersection, the crash partner continued straight into the intersection against the traffic control and did not yield. The ego vehicle applied braking but was unable to avoid contact, resulting in a side-impact collision within the intersection.
\end{tcolorbox}

\begin{tcolorbox}[colback=gray!10, colframe=gray!50, boxrule=0.5pt]

\textbf{Cluster-10.}

Road: Intersection, SV: Making Left Turn, CP: Proceeding Straight, Work Zone: No 

\textbf{Narrative generated.}

Ego vehicle approached a signalized intersection in the left-turn lane and initiated a left turn on a permissive phase. A crash-partner vehicle was proceeding straight through the intersection from the cross street. As ego entered the intersection to complete the turn, the crash-partner continued straight at a steady speed without yielding, and the vehicles converged in the center of the intersection, resulting in a side-impact collision as ego crossed the crash-partner’s path.
\end{tcolorbox}

\begin{tcolorbox}[colback=gray!10, colframe=gray!50, boxrule=0.5pt]
\textbf{Cluster-11.}

Road: Highway/Freeway, SV: Proceeding Straight, CP: Proceeding Straight, Work Zone: No 

\textbf{Narrative generated.}

On [XXX], 2022, the ego vehicle was traveling eastbound on a limited-access freeway, proceeding straight in the right travel lane at a steady speed. A crash partner passenger vehicle was proceeding straight in the adjacent lane to the left, traveling parallel to the ego vehicle. A short line of traffic cones had been placed along the dashed lane line ahead (not an active work zone), narrowing the usable space between the two lanes. As both vehicles continued straight, the crash partner tracked slightly right of center in its lane while passing the cones, encroaching across the lane line. The crash partner’s right side contacted the ego vehicle’s left front/left side, resulting in a sideswipe-type collision while both vehicles were proceeding straight.

\end{tcolorbox}

\begin{tcolorbox}[colback=gray!10, colframe=gray!50, boxrule=0.5pt]
\textbf{Cluster-12.}

Road: Highway/Freeway, SV: Proceeding Straight, CP: Proceeding Straight, Work Zone: Yes 

\textbf{Narrative generated.}

On April [XXX] at approximately 10:35 AM CT, the ego vehicle was traveling eastbound on a highway through an active work zone, proceeding straight in the right through-lane at a steady speed. A crash-partner SUV was traveling in the adjacent left through-lane, also proceeding straight at a similar speed. The work zone used cones to taper and narrow both lanes, reducing lateral clearance between vehicles while maintaining straight-ahead travel. As both vehicles entered the narrowed section, the crash-partner SUV tracked slightly toward the right within its lane while continuing forward, and its right side encroached into the ego vehicle’s lane space. The ego vehicle and the crash-partner SUV made contact in a sideswipe along their left/right sides, respectively, while both continued proceeding straight through the coned work zone.

\end{tcolorbox}

\begin{tcolorbox}[colback=gray!10, colframe=gray!50, boxrule=0.5pt]
\textbf{Cluster-13.}

Road: Highway/Freeway, SV: Proceeding Straight, CP: Proceeding Straight, Work Zone: Yes 

\textbf{Narrative generated.}

On [XXX], 2024 at [XXX] PT, an autonomous vehicle was traveling proceeding straight on a highway through an active work zone with cones tapering and narrowing the available lanes. A passenger vehicle traveling proceeding straight in the adjacent lane approached the cone taper and continued forward as the lane boundary shifted, gradually encroaching toward the vehicle’s lane. The vehicle reduced speed to maintain a safe buffer in response to the narrowing roadway, but the passenger vehicle’s continued forward motion alongside the taper resulted in a sideswipe contact along the vehicle’s left side as both vehicles remained traveling straight through the work zone.

\end{tcolorbox}

\begin{tcolorbox}[colback=gray!10, colframe=gray!50, boxrule=0.5pt]
\textbf{Cluster-14.}

Road: Intersection, SV: Proceeding Straight, CP: Making Left Turn, Work Zone: No 

\textbf{Narrative generated.}

On Tuesday, February [XXX], 2025 at 3:18 PM local time, the ego vehicle was proceeding straight through a signalized intersection at a steady speed. A crash partner vehicle was positioned on the cross street at the intersection and initiated a left turn to enter and merge onto the ego vehicle’s roadway. As the crash partner accelerated into the turn, it entered the ego vehicle’s lane while the ego vehicle was already committed to the intersection. The ego vehicle applied braking but was unable to avoid contact. The ego vehicle’s front-left area contacted the crash partner’s right-side area within the intersection as the crash partner continued the left-turn path. No work zone was present.

\end{tcolorbox}

\begin{tcolorbox}[colback=gray!10, colframe=gray!50, boxrule=0.5pt]
\textbf{Cluster-15.}

Road: Intersection, SV: Making Right Turn, CP: Making Right Turn, Work Zone: No 

\textbf{Narrative generated.}

A vehicle (Vehicle 1) operating in autonomy approached a signalized intersection in the inside of two dedicated right‑turn lanes and began a right turn on a green indication. A passenger vehicle (Vehicle 2) in the adjacent outside right‑turn lane also initiated a right turn at approximately the same time. During the turn, Vehicle 2 tracked wider than its marked lane and encroached into the inside receiving lane, moving laterally into Vehicle 1’s path. Vehicle 1 braked to maintain a safe gap but was unable to fully avoid contact, resulting in a low‑speed sideswipe/angle impact within the intersection as both vehicles completed their right turns. No work zone was present.

\end{tcolorbox}

\begin{tcolorbox}[colback=gray!10, colframe=gray!50, boxrule=0.5pt]
\textbf{Cluster-16.}

Road: Intersection, SV: Proceeding Straight, CP: Proceeding Straight, Work Zone: No 

\textbf{Narrative generated.}

On March [XXX], 2025 at 6:18 PM local time, an autonomous vehicle (AV) was proceeding straight through a signalized intersection in its lane at a steady target speed. A crash partner vehicle approached the same intersection from the cross street and also proceeded straight. As the AV entered the intersection, the crash partner continued straight into the intersection at a higher speed than the available gap and did not yield, resulting in a side-impact collision within the intersection (front of the crash partner contacting the AV’s right side). No work zone was present.

\end{tcolorbox}

\begin{tcolorbox}[colback=gray!10, colframe=gray!50, boxrule=0.5pt]
\textbf{Cluster-17.}

Road: Intersection, SV: Proceeding Straight, CP: Proceeding Straight, Work Zone: No 

\textbf{Narrative generated.}

A self-driving vehicle (“Ego AV”) was proceeding straight through an intersection at a steady speed. A crash partner vehicle was also proceeding straight on the crossing street and approached the same intersection. As the Ego AV entered the intersection, the crash partner vehicle accelerated into the intersection and then braked late, reducing the available gap. The Ego AV applied braking to maintain a safe stopping distance, but the vehicles entered the conflict point at the same time and collided in the intersection (side-impact angle), with neither vehicle striking the other from behind and without a head-on approach.
\end{tcolorbox}

\begin{tcolorbox}[colback=gray!10, colframe=gray!50, boxrule=0.5pt]
\textbf{Cluster-18.}

Road: Intersection, SV: Making Left Turn, CP: Proceeding Straight, Work Zone: No 

\textbf{Narrative generated.}

On Tuesday, March [XXX], 2025 at 6:12 PM local time, the ego vehicle was traveling straight through a signalized four-way intersection in the right through lane at approximately 23 mph with a green indication. A passenger car on the cross street approached the intersection from the ego vehicle’s right side, slowed, then initiated a left turn to enter the ego vehicle’s roadway (turning to travel in the same direction as the ego vehicle). As the car turned, it accelerated into the ego vehicle’s lane and occupied the lane space ahead of the ego vehicle before completing the turn. The ego vehicle applied braking but the available gap was insufficient, and the front of the ego vehicle contacted the car’s right-rear quarter panel within the intersection. The vehicles came to rest shortly after the impact; no work zone activity was present.

\end{tcolorbox}

\begin{tcolorbox}[colback=gray!10, colframe=gray!50, boxrule=0.5pt]
\textbf{Cluster-19.}

Road: Intersection, SV: Making Left Turn, CP:Making Left Turn, Work Zone: No 

\textbf{Narrative generated.}

On Tuesday, April [XXX], 2025 at 6:12 PM local time, the ego vehicle was proceeding straight through a signalized intersection in the right-most through lane at approximately 25 mph under a green indication. A crash partner vehicle approached the same intersection from the cross street and initiated a left turn to enter the ego vehicle’s roadway. As the crash partner turned, it entered the ego vehicle’s lane while the ego vehicle was already committed to the intersection. The ego vehicle applied braking and reduced speed but was unable to avoid contact, and the front of the ego vehicle struck the right side of the crash partner within the intersection. No work zone was present.
\end{tcolorbox}


\begin{table}[h]
\centering
\setlength{\tabcolsep}{4pt}
\renewcommand{\arraystretch}{1.1}

\begin{tabular}{p{0.24\columnwidth} p{0.26\columnwidth} p{0.42\columnwidth}}
\toprule

\textbf{Variable} & \textbf{Permissible Values} & \textbf{Description} \\
\midrule

Map Type 
& \textsf{I}, \textsf{T}, \textsf{O}, \textsf{C}, \textsf{S} 
& Reflects road-type based on meta variable. \\

Target Speed (Self-Vehicle) 
& Continuous range (m/s) 
& Adopted from narrative and tuned to suit the scenario. \\

Traffic Density 
& $[0,1]$ 
& Controls density of randomized background traffic. \\

\midrule
\multicolumn{3}{c}{\textbf{Crash Partner (CP) Specific Variables}} \\
\midrule

Lane Offset 
& $\{-1,0,+1\}$ 
& Initialization of crash partner lane relative to ego lane. \\

$\Delta_s$, $\Delta_t$ 
& Longitudinal / lateral offset (m) 
& Relative position of non-ego crash partner. \\


Policy 
& \textsf{Sudden}, \textsf{Accelerate-then-brake}, \textsf{IDM} 
& CP control policy: sudden braking, accelerate-then-brake, or IDM. Includes policy-specific parameters such as brake timing and magnitude. \\

\bottomrule
\end{tabular}

\caption{Scenario variables and permissible values used for LLM-based parameterization.}
\label{tab:scene-var}
\end{table}

\section{Prompts}\label{app:prompts}
\subsection{Paraphrasing agent prompt}\label{app:prompt-1}

\begin{tcolorbox}[
  title={Scenario Synthesis from Historical Crash Records},
  colback=white,
  colframe=black!60,
  boxrule=0.4pt,
  arc=2pt,
  left=6pt,right=6pt,top=6pt,bottom=6pt
]

Shown below are failure scenarios from actual historical crash records. Use these as examples of realistic crashes and propose a new scenario that represents a scenario with the following attributes:

\vspace{4pt}
\textbf{Required attributes}
\begin{itemize}\setlength\itemsep{2pt}
  \item \textbf{Pre-Crash movement (Crash Partner)}
  \item \textbf{Pre-Crash movement (Self Vehicle)}
  \item \textbf{Road Type}
  \item \textbf{Work zone}
  \item \textbf{Narrative lists}
\end{itemize}

\vspace{4pt}
\textbf{Constraints (do not generate)}
\begin{itemize}\setlength\itemsep{2pt}
  \item Do not design scenarios that are rear-end collisions for the ego vehicle, or where the ego vehicle is hit by a non-ego vehicle from behind, because we would like to test the ego vehicle's policy for braking at a safe distance.
  \item Do not design scenarios with a head-on collision between ego and non-ego vehicle coming at each other from opposite directions.
  \item For example, avoid narratives such as: ``An oncoming passenger vehicle in the opposite direction entered the intersection and initiated a left turn across the vehicle’s path, appearing to misjudge the vehicle’s speed and gap.''
\end{itemize}

\vspace{4pt}
\textbf{Recreation constraints (capabilities)}
\begin{itemize}\setlength\itemsep{2pt}
  \item We can control spawn positions of ego and non-ego vehicles.
  \item We can scatter cones.
  \item We can control target speeds of each vehicle.
  \item We can control movement of non-ego vehicle as \texttt{sudden\_brake}, \texttt{accel\_then\_brake}, or nominal path following.
  \item Design a scenario that can be recreated using these variables; do not add additional details.
  \item At present, sharp changes in trajectory curvature and direction of motion of vehicles cannot be adjusted.
\end{itemize}

\vspace{4pt}
\textbf{Output format}: Return a natural language output (\texttt{text/str}) that mimics a narrative.

\end{tcolorbox}

\subsection{Scenario design agent prompt}\label{app:prompt-2}
\begin{figure}[h]
\centering
\footnotesize
\begin{tcolorbox}[
  colback=white,
  colframe=black!70,
  boxrule=0.4pt,
  arc=2pt,
  left=4pt,right=4pt,top=4pt,bottom=4pt,
  width=\linewidth
]

\textbf{MetaDrive Scenario JSON Generation Prompt}

\textbf{Goal:} Generate a MetaDrive scenario JSON for potential collision/failure cases involving an ego vehicle (IDM policy) and crash partner(s) (CP/NCP).

\textbf{Constraint:} Ego uses IDM (collision-avoidant). Create simple but realistic failures observable under IDM (e.g., early stopping, phantom braking, oscillation). Avoid naive crashes.

\textbf{Crash Scenario Desired:} \{LLM-generated Narrative\}

\textbf{Reference JSON Examples:} \{examples\}

\vspace{3pt}
\textbf{Map Types:}
\texttt{'X'} (intersection), 
\texttt{'O'} (circle), 
\texttt{'T'} (3-way), 
\texttt{'S'} (straight), 
\texttt{'C'} (curved).
Choose based on narrative.

\vspace{3pt}
\textbf{Turn Specification:}
\texttt{ego\_turn: 'left'|'right'|'straight'} \\
\texttt{lead\_turn: 'left'|'right'|'straight'}

\vspace{3pt}
\textbf{Each NCP must include:}
\begin{itemize}\itemsep1pt
\item \texttt{id}
\item \texttt{direction: "same"|"opposite"}
\item \texttt{lane\_offset: 0,$\pm1$}
\item \texttt{delta\_s\_m} (longitudinal; negative = behind)
\item \texttt{delta\_t\_m} (lateral)
\item \texttt{policy: "sudden\_brake"|"accel\_then\_brake"|"idm\_nominal"}
\item \texttt{params} (policy-specific dict)
\end{itemize}

\vspace{3pt}
\textbf{Policy Params:}

\texttt{sudden\_brake:} 
\{target\_speed, brake\_step, brake\_steps, speed\_kp(optional)\}

\texttt{accel\_then\_brake:}
\{accel\_mag, accel\_steps, cruise\_speed, brake\_step, brake\_steps, speed\_kp(optional)\}

\texttt{idm\_nominal:}
\{target\_speed\_mps\}

\vspace{3pt}
\textbf{Static Obstacles:} Only include cones if work zone = Yes or narrative specifies obstacles.

\vspace{3pt}
\textbf{Validity:}
Follow example JSON format exactly. Avoid incomplete dicts or invalid lane configurations (e.g., navigation errors).

\end{tcolorbox}
\caption{Prompt used by scenario generation agent for actual schema generation.}
\end{figure}

\subsection{Fine tuning agent prompt}\label{app:prompt-3}

\begin{tcolorbox}[
  title={Scenario Parameter Optimization and Fine-Tuning},
  colback=white,
  colframe=black!60,
  boxrule=0.4pt,
  arc=2pt,
  left=6pt,right=6pt,top=6pt,bottom=6pt
]

You are being asked to optimize the scenario parameters. Analyze the scenario dict, recreate the scenario spatially, and think what changes can be made to parameters such as speed, positioning, etc. to induce a collision/crash/accident/unexpected behavior such as abrupt braking, oscillation, etc.

\vspace{4pt}
Make sure that tweaking parameters does not violate the narrative provided, as the spatial arrangement and scene playout must follow the narrative.

\vspace{6pt}
\textbf{Inputs provided}
\begin{itemize}\setlength\itemsep{2pt}
  \item \textbf{Original scenario dict}
  \item \textbf{Narrative}
  \item \textbf{Metrics for the scenario dict}: Minimum Euclidean gap, Time at minimum Euclidean gap
\end{itemize}

\vspace{4pt}
\textbf{How to use the metrics}
\begin{itemize}\setlength\itemsep{2pt}
  \item Use these metrics to adjust scenario parameters, e.g., changing speeds, longitudinal gap, or lateral gap at initialization.
  \item Minimum Euclidean gap must be less than 2 meters to be considered a collision.
  \item Use the time of minimum gap to determine if it is consistent with the narrative (e.g., the narrative suggests collision much sooner than the current scene).
\end{itemize}

\vspace{4pt}
\textbf{Avoid degenerate solutions}
\begin{itemize}\setlength\itemsep{2pt}
  \item Avoid tweaking quantities such that the minimum gap occurs naively at the beginning or end of the trajectory.
  \item For example, avoid collisions at timestep 0 or 1 (very early) or timestep 359 or 412 (very late), i.e., at the very beginning or end of the trajectory.
\end{itemize}

\vspace{4pt}
\textbf{Output format}
\begin{itemize}\setlength\itemsep{2pt}
  \item Return a revised scenario in JSON format that only performs fine-tuning to the original DSL.
  \item Only change the original DSL substantially if it is violating the narrative; otherwise stick to numerical fine-tuning.
\end{itemize}

\end{tcolorbox}

\end{document}